\documentclass[twocolumn,showpacs,%
  nofootinbib,aps,superscriptaddress,%
  eqsecnum,prd,notitlepage,showkeys,10pt]{revtex4-1}

\usepackage{amssymb}
\usepackage{amsmath}
\usepackage{graphicx}
\usepackage{dcolumn}
\usepackage{hyperref}
\usepackage{xcolor}

\begin{document}

\title{Sleep-Like Unsupervised Replay Improves Performance when Data are Limited or Unbalanced}

\affiliation{Del Norte High School, 16601 Nighthawk Ln, San Diego, CA 92127}
\affiliation{Department of Medicine, University of California, San Diego}
\affiliation{Department of Computer Science \& Engineering, University of California, San Diego\\ 
    9500 Gilman Drive, La Jolla, CA 92092}
    
\author{Anthony Bazhenov$^{1,*}$}
\author{\textcolor{white}{-} Pahan Dewasurendra$^{1,*}$}
\author{\textcolor{white}{-} Giri Krishnan$^{2}$ }
\author{\textcolor{white}{-} Jean Erik Delanois$^{2,3}$ }

\maketitle

\def\thefootnote{*}\footnotetext{These authors contributed equally}\def\thefootnote{\arabic{footnote}}

\section*{Abstract}

 The performance of artificial neural networks (ANNs) degrades when training data are limited or imbalanced. In contrast, the human brain can learn quickly from just a few examples. Here, we investigated the role of sleep in improving the performance of ANNs trained with limited data on the MNIST and Fashion MNIST datasets. Sleep was implemented as an unsupervised phase with local Hebbian type learning rules. We found a significant boost in accuracy after the sleep phase for models trained with limited data in the range of 0.5-10\% of total MNIST or Fashion MNIST datasets. When more than 10\% of the total data was used, sleep alone had a slight negative impact on performance, but this was remedied by fine-tuning on the original data. This study sheds light on a potential synaptic weight dynamics strategy employed by the brain during sleep to enhance memory performance when training data are limited or imbalanced.

%\textbf{TODO - Erik will add template for 2024 (currently it is 2023 template}
\section{Introduction}

Deep learning methods have shown considerable performance when training datasets are large, however, existing techniques generally fail in low training data conditions. Additionally, training datasets are often imbalanced, with some categories occurring more frequently than others, resulting in reduced accuracy for ANNs. Several methods have been proposed to overcome these limitations. These include data augmentation \cite{shorten2019survey}, pre-training on other datasets \cite{zhuang2020comprehensive} or alternative architectures such as neural tangent kernel \cite{arora2019harnessing}. However, these approaches do not address the fundamental question of how to make overparameterized deep learning networks learn to generalize from small datasets without overfitting. 
%Our study proposes an approach to address this problem.
In contrast, the human brain demonstrates the ability to learn quickly from just a few examples. %It has been suggested that memory replay during biological sleep can strengthen memories learned during wakefulness.

Sleep has been shown to play an important role in memory consolidation in biological systems \cite{stickgold2005sleep}.
%, lewis2011overlapping}. %During sleep, neurons are spontaneously active without external input and generate complex patterns of synchronized activity across brain regions \cite{steriade1993thalamocortical, krishnan2016cellular}. 
Two critical components which are believed to underlie memory consolidation during sleep are spontaneous replay of memory traces and local unsupervised synaptic plasticity that  restricts synaptic changes to relevant memories only. %\cite{wilson1994reactivation, stickgold2005sleep}. 
%, wei2016synaptic}.  
During sleep, replay of recently learned memories along with relevant old memories 
%,lewis2018memory, hennevin1995processing, rasch2013sleep, mednick2011opportunistic, paller2004memory, oudiette2013role}
enables the network to form stable long-term memory representations \cite{rasch2013sleep} and reduces competition between memories
%orthogonal memory representations to enable coexistence of competing memories within overlapping populations of neurons 
\cite{
%wei2016synaptic, 
gonzalez2020can, golden2020sleep}. The idea of replay has been explored in machine learning to enable continual learning. However, spontaneous unsupervised replay found in the biological brain and implemented here is significantly different compared to explicit replay of past inputs implemented in machine learning rehearsal methods \cite{hayes2021replay}.

%Unlike the standard global optimization used in machine learning (such as backpropagation), local plasticity allows synaptic changes to affect only relevant memories. 

%While consolidation of declarative memories presumably depends on the interplay between fast-learning hippocampus and slow-learning cortex \cite{rasch2013sleep} ('Complementary Learning Systems Theory' \cite{mcclelland1995there}), several types of procedural memories (e.g., skills) are believed to be hippocampus-independent and still require consolidation during sleep, particulartly during Rapid Eye Movement (REM) sleep  \cite{mcdevitt2015rem}. 
These results from neuroscience suggest that applying sleep replay principles to ANNs may enhance memory representations and, consequently, improve the performance of machine learning models trained on limited or unbalanced datasets, as tested in our study. 

%We show that implementing a sleep-like phase after an ANN learns a new task enables replay and makes possible continual learning of multiple tasks without forgetting. These results are formalized as a Sleep Replay Consolidation  (SRC) algorithm as follows. First, an ANN is trained using the backpropagation algorithm, denoted below as awake training. Next, spontaneous brain dynamics, similar to those found in sleep \cite{krishnan2016cellular, bazhenov2002model}, are simulated and we run one-time step of network simulation propagating spontaneous activity forward through the network. Next, we do a backward pass through the network in order to apply local Hebbian plasticity rules to modify weights. After running multiple steps of this unsupervised training phase, testing or further training using regular backpropagation is performed. SRC can be applied alone or combined with state-of-the-art rehearsal methods to further improve these methods' performance. We recently found that the same approach can promote domain generalization and improve robustness against adversarial attacks \cite{tadros2019}. Here, we develop the algorithm to continual learning problem by limiting the amount of information available during sleep. We show that spontaneous reactivation of neurons during sleep engages local plasticity rules that can recover performance on tasks that were thought to be lost due to catastrophic forgetting after new task training.

\section{Algorithm}
 A fully-connected ANN with two hidden layers was first trained on a randomly selected subset of MNIST or Fashion MNIST (FMNIST) datasets using backpropagation. Subsequently, the sleep replay consolidation (SRC) algorithm was implemented as previously described in \cite{Tadros_NC2022}. Briefly (see Supplementary Material for details), the ANN trained by limited data was mapped to a spiking neural network (SNN) with the same architecture. %During the sleep phase, 
 The SNN's activity was driven by randomly distributed Poisson spiking input that reflected average inputs observed in the training dataset. Local Hebbian-type plasticity was implemented to modify weights during the sleep phase, i.e., synaptic strength was increased if presynaptic activation was followed by postsynaptic activation and reduced if postsynaptic activation occurred without presynaptic activation. After the sleep phase, the SNN was remapped back to an ANN. In \cite{Tadros_NC2022} SRC was applied after each new task training to avoid catastrophic forgetting, here we applied it once after training with limited data. %Finally performance was tested again immediately or after further fine-tuning on the original limited dataset.

\section{Results}
\begin{figure}[t!]
\centering
\includegraphics[width=0.9\columnwidth]{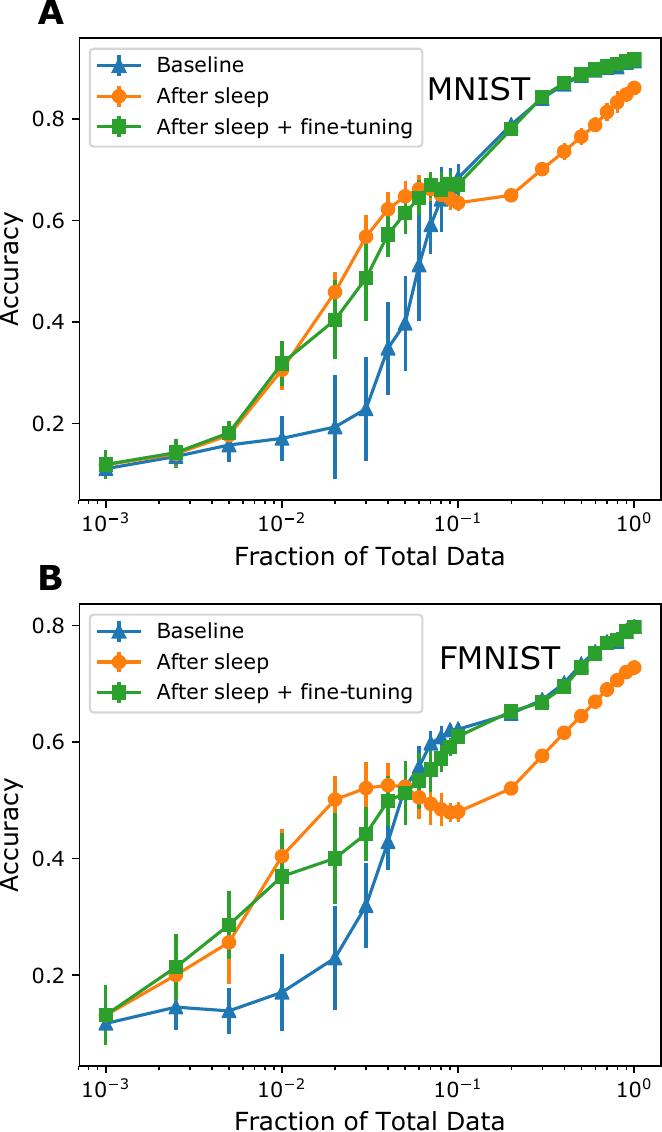} % Reduce the figure size so that it is slightly narrower than the column. Don't use precise values for figure width.This setup will avoid overfull boxes.
\caption{Accuracy on MNIST \textbf{(A)} and FMNIST \textbf{(B)} with mean (lines) and standard deviation (error bars) across 10 trials. X-axis - log of the relative amount of data used for training (e.g., 0.01=1\% of data). Blue - baseline (after ANN training); Orange - baseline + sleep; Green - baseline + sleep + fine-tuning. Note significant gain in accuracy after sleep phase on low data. The sleep phase reduced performance on high data but was largely recovered by fine-tuning.
}

% Accuracy on MNIST \textbf{(A)} and Fashion MNIST \textbf{(B)} data set. X-axis - log of the relative amount of data used for training (e.g., 0.01=1\% of data). Blue line - baseline accuracy after training; Orange line - accuracy after baseline training followed by sleep; Green line - accuracy after sleep followed by fine-tuning by original data. Note significant gain in accuracy after sleep phase in the range 0.5-8\% / 0.2-4\% for MNIST / FMNIST of data. For $>$10\% of data, sleep phase reduced performance but it was largely recovered by fine-tuning.

% (A) MNIST data. Note significant gain in accuracy after sleep phase in the range 0.5-8\% of data. For $>$10\% of data, sleep phase reduced performance but it was largely recovered by fine-tuning. (B) Fashion MNIST. Note gain in accuracy in the range 0.2-4\% of data.
%}
\label{fig:fig1}
\end{figure}

%In this study, we investigated the role of sleep and its impact on improving the performance of ANNs trained with limited data. Initially, the ANN was trained on the MNIST or Fashion MNIST datasets using backpropagation and subsequently mapped to a spiking neural network (SNN) with the same architecture, incorporating sleep. During the sleep phase, the SNN's activity was driven by randomly distributed Poisson spiking input, and synaptic weights were updated based on local Hebbian-type plasticity rules (See Supplemenets for details). After the sleep phase, the SNN was remapped back to the ANN, and its accuracy was evaluated again. 
When the ANN was trained with the full dataset, it achieved an accuracy of over 90\%. However, when less than 10\% of the data was used during training, accuracy significantly declined (Figure \ref{fig:fig1}, blue line). When  0.5\% to 10\% of the total data was used for ANN training, the subsequent application of SRC resulted in a substantial (20-30\%) increase in accuracy for both MNIST and Fashion MNIST datasets (Figure \ref{fig:fig1}, orange line). Increasing the training duration (number of epochs), increased performance before sleep but a significant performance gain after sleep remained. 

Analysis of the confusion matrix (see Supplementary Material) revealed that networks trained with limited data can exhibit biases towards a few classes. For example, when 3\% of the MNIST data was used in training, classes 0, 2, 5, and 6 were all classified as 0. However, after sleep, classes 0, 2, and 6 were classified correctly. Succinctly, the model exhibited a more balanced response after the application of SRC.
%Confusion matrix analysis (see Supplementary Material), revealed that in such training conditions only few classes were predicted with a reasonable accuracy while others were  classified incorrectly, usually as a single digit (e.g., classes 3-9 were all classified as  3 for MNIST). 

%We found that when only 2-10\% of the data was used for ANN training, SRC resulted in a substantial (20-30\%) increase in accuracy for both MNIST and Fashion MNIST datasets (see Figure \ref{fig:fig1}). 

%\newpage

%\begin{figure}[t]
%\centering
%\includegraphics[width=0.9\columnwidth]{fmnist-mean-std-log.png} % Reduce the figure size so that it is slightly narrower than the column. Don't use precise values for figure width.This setup will avoid overfull boxes.
%\caption{FMNIST plot  (add labels, legend)}
%\label{fig:fig2}
%\end{figure}

While performance improved when there was limited training data, we also observed a slight (10-15\%) decrease in performance when more than approximately 10\% of the data was employed for ANN training. We found that this decrease in performance could be mitigated by fine-tuning the ANN after sleep using the original (limited) training data (Figure \ref{fig:fig1}, green line). Thus, by incorporating both sleep and fine-tuning, we were able to maintain performance on models trained with the full dataset while still achieving performance gains on models trained with limited data. 

Next, we examined accuracy when a significant class imbalance was introduced to the training set by selectively reducing the number of training examples used for certain classes. We found that class-wise model performance was more robust to data reduction for some classes when compared to
others. After SRC, most classes
showed a positive improvement in class-wise accuracy (see Supplementary Material).
Thus, the sleep phase proved effective in increasing model accuracy on underrepresented classes while preserving accuracy on well-trained classes. 
%Analysis of the confusion matrix revealed that networks trained with limited data can exhibit biases towards a few classes, whereas after sleep, they demonstrated a more balanced response. 

Analysis of synaptic weights revealed that  SRC increased strength for a small fraction of critical synapses, while many other synapses were weakened (see Supplementary Material). This suggests that the overall accuracy increase after SRC was a result of increasing the sparsity of responses.
%Overall, sleep improved the accuracy by increasing the sparsity of responses. resulted from reducing synaptic weights. 

Our study sheds light on a potential synaptic weight dynamics strategy employed by the brain during sleep to enhance memory performance when training data are limited or imbalanced. Applied to ANNs, sleep-like replay improves performance in a completely unsupervised manner, requiring no additional data, and can be applied to already trained models. 
\vspace{-10 pt}
\section{Acknowledgments}
\vspace{-5 pt}
Supported by NSF (grants 2209874 and 2223839).
%\textbf{TODO - placeholder short space for Acknowledgments - to be inserted after review}
\vspace{-10 pt}
% \bibentry{c:23}.

% ARXIV SUBMISSION NEEDS THIS COMMENTED OUT
\bibliography{bib}

%merlin.mbs apsrev4-1.bst 2010-07-25 4.21a (PWD, AO, DPC) hacked
%Control: key (0)
%Control: author (8) initials jnrlst
%Control: editor formatted (1) identically to author
%Control: production of article title (-1) disabled
%Control: page (0) single
%Control: year (1) truncated
%Control: production of eprint (0) enabled
\begin{thebibliography}{9}%
\makeatletter
\providecommand \@ifxundefined [1]{%
 \@ifx{#1\undefined}
}%
\providecommand \@ifnum [1]{%
 \ifnum #1\expandafter \@firstoftwo
 \else \expandafter \@secondoftwo
 \fi
}%
\providecommand \@ifx [1]{%
 \ifx #1\expandafter \@firstoftwo
 \else \expandafter \@secondoftwo
 \fi
}%
\providecommand \natexlab [1]{#1}%
\providecommand \enquote  [1]{``#1''}%
\providecommand \bibnamefont  [1]{#1}%
\providecommand \bibfnamefont [1]{#1}%
\providecommand \citenamefont [1]{#1}%
\providecommand \href@noop [0]{\@secondoftwo}%
\providecommand \href [0]{\begingroup \@sanitize@url \@href}%
\providecommand \@href[1]{\@@startlink{#1}\@@href}%
\providecommand \@@href[1]{\endgroup#1\@@endlink}%
\providecommand \@sanitize@url [0]{\catcode `\\12\catcode `\$12\catcode `\&12\catcode `\#12\catcode `\^12\catcode `\_12\catcode `\%12\relax}%
\providecommand \@@startlink[1]{}%
\providecommand \@@endlink[0]{}%
\providecommand \url  [0]{\begingroup\@sanitize@url \@url }%
\providecommand \@url [1]{\endgroup\@href {#1}{\urlprefix }}%
\providecommand \urlprefix  [0]{URL }%
\providecommand \Eprint [0]{\href }%
\providecommand \doibase [0]{http://dx.doi.org/}%
\providecommand \selectlanguage [0]{\@gobble}%
\providecommand \bibinfo  [0]{\@secondoftwo}%
\providecommand \bibfield  [0]{\@secondoftwo}%
\providecommand \translation [1]{[#1]}%
\providecommand \BibitemOpen [0]{}%
\providecommand \bibitemStop [0]{}%
\providecommand \bibitemNoStop [0]{.\EOS\space}%
\providecommand \EOS [0]{\spacefactor3000\relax}%
\providecommand \BibitemShut  [1]{\csname bibitem#1\endcsname}%
\let\auto@bib@innerbib\@empty
%</preamble>
\bibitem [{\citenamefont {Shorten}\ and\ \citenamefont {Khoshgoftaar}(2019)}]{shorten2019survey}%
  \BibitemOpen
  \bibfield  {author} {\bibinfo {author} {\bibfnamefont {C.}~\bibnamefont {Shorten}}\ and\ \bibinfo {author} {\bibfnamefont {T.~M.}\ \bibnamefont {Khoshgoftaar}},\ }\href@noop {} {\bibfield  {journal} {\bibinfo  {journal} {Journal of big data}\ }\textbf {\bibinfo {volume} {6}},\ \bibinfo {pages} {1} (\bibinfo {year} {2019})}\BibitemShut {NoStop}%
\bibitem [{\citenamefont {Zhuang}\ \emph {et~al.}(2020)\citenamefont {Zhuang}, \citenamefont {Qi}, \citenamefont {Duan}, \citenamefont {Xi}, \citenamefont {Zhu}, \citenamefont {Zhu}, \citenamefont {Xiong},\ and\ \citenamefont {He}}]{zhuang2020comprehensive}%
  \BibitemOpen
  \bibfield  {author} {\bibinfo {author} {\bibfnamefont {F.}~\bibnamefont {Zhuang}}, \bibinfo {author} {\bibfnamefont {Z.}~\bibnamefont {Qi}}, \bibinfo {author} {\bibfnamefont {K.}~\bibnamefont {Duan}}, \bibinfo {author} {\bibfnamefont {D.}~\bibnamefont {Xi}}, \bibinfo {author} {\bibfnamefont {Y.}~\bibnamefont {Zhu}}, \bibinfo {author} {\bibfnamefont {H.}~\bibnamefont {Zhu}}, \bibinfo {author} {\bibfnamefont {H.}~\bibnamefont {Xiong}}, \ and\ \bibinfo {author} {\bibfnamefont {Q.}~\bibnamefont {He}},\ }\href@noop {} {\bibfield  {journal} {\bibinfo  {journal} {Proceedings of the IEEE}\ }\textbf {\bibinfo {volume} {109}},\ \bibinfo {pages} {43} (\bibinfo {year} {2020})}\BibitemShut {NoStop}%
\bibitem [{\citenamefont {Arora}\ \emph {et~al.}(2019)\citenamefont {Arora}, \citenamefont {Du}, \citenamefont {Li}, \citenamefont {Salakhutdinov}, \citenamefont {Wang},\ and\ \citenamefont {Yu}}]{arora2019harnessing}%
  \BibitemOpen
  \bibfield  {author} {\bibinfo {author} {\bibfnamefont {S.}~\bibnamefont {Arora}}, \bibinfo {author} {\bibfnamefont {S.~S.}\ \bibnamefont {Du}}, \bibinfo {author} {\bibfnamefont {Z.}~\bibnamefont {Li}}, \bibinfo {author} {\bibfnamefont {R.}~\bibnamefont {Salakhutdinov}}, \bibinfo {author} {\bibfnamefont {R.}~\bibnamefont {Wang}}, \ and\ \bibinfo {author} {\bibfnamefont {D.}~\bibnamefont {Yu}},\ }\href@noop {} {\bibfield  {journal} {\bibinfo  {journal} {arXiv preprint arXiv:1910.01663}\ } (\bibinfo {year} {2019})}\BibitemShut {NoStop}%
\bibitem [{\citenamefont {Stickgold}(2005)}]{stickgold2005sleep}%
  \BibitemOpen
  \bibfield  {author} {\bibinfo {author} {\bibfnamefont {R.}~\bibnamefont {Stickgold}},\ }\href@noop {} {\bibfield  {journal} {\bibinfo  {journal} {Nature}\ }\textbf {\bibinfo {volume} {437}},\ \bibinfo {pages} {1272} (\bibinfo {year} {2005})}\BibitemShut {NoStop}%
\bibitem [{\citenamefont {Rasch}\ and\ \citenamefont {Born}(2013)}]{rasch2013sleep}%
  \BibitemOpen
  \bibfield  {author} {\bibinfo {author} {\bibfnamefont {B.}~\bibnamefont {Rasch}}\ and\ \bibinfo {author} {\bibfnamefont {J.}~\bibnamefont {Born}},\ }\href@noop {} {\bibfield  {journal} {\bibinfo  {journal} {Physiological reviews}\ }\textbf {\bibinfo {volume} {93}},\ \bibinfo {pages} {681} (\bibinfo {year} {2013})}\BibitemShut {NoStop}%
\bibitem [{\citenamefont {Gonz{\'a}lez}\ \emph {et~al.}(2020)\citenamefont {Gonz{\'a}lez}, \citenamefont {Sokolov}, \citenamefont {Krishnan}, \citenamefont {Delanois},\ and\ \citenamefont {Bazhenov}}]{gonzalez2020can}%
  \BibitemOpen
  \bibfield  {author} {\bibinfo {author} {\bibfnamefont {O.~C.}\ \bibnamefont {Gonz{\'a}lez}}, \bibinfo {author} {\bibfnamefont {Y.}~\bibnamefont {Sokolov}}, \bibinfo {author} {\bibfnamefont {G.~P.}\ \bibnamefont {Krishnan}}, \bibinfo {author} {\bibfnamefont {J.~E.}\ \bibnamefont {Delanois}}, \ and\ \bibinfo {author} {\bibfnamefont {M.}~\bibnamefont {Bazhenov}},\ }\href@noop {} {\bibfield  {journal} {\bibinfo  {journal} {Elife}\ }\textbf {\bibinfo {volume} {9}},\ \bibinfo {pages} {e51005} (\bibinfo {year} {2020})}\BibitemShut {NoStop}%
\bibitem [{\citenamefont {Golden}\ \emph {et~al.}(2022)\citenamefont {Golden}, \citenamefont {Delanois}, \citenamefont {Sanda},\ and\ \citenamefont {Bazhenov}}]{golden2020sleep}%
  \BibitemOpen
  \bibfield  {author} {\bibinfo {author} {\bibfnamefont {R.}~\bibnamefont {Golden}}, \bibinfo {author} {\bibfnamefont {J.~E.}\ \bibnamefont {Delanois}}, \bibinfo {author} {\bibfnamefont {P.}~\bibnamefont {Sanda}}, \ and\ \bibinfo {author} {\bibfnamefont {M.}~\bibnamefont {Bazhenov}},\ }\href@noop {} {\bibfield  {journal} {\bibinfo  {journal} {PLoS Computational Biology}\ }\textbf {\bibinfo {volume} {18}},\ \bibinfo {pages} {e1010628} (\bibinfo {year} {2022})}\BibitemShut {NoStop}%
\bibitem [{\citenamefont {Hayes}\ \emph {et~al.}(2021)\citenamefont {Hayes}, \citenamefont {Krishnan}, \citenamefont {Bazhenov}, \citenamefont {Siegelmann}, \citenamefont {Sejnowski},\ and\ \citenamefont {Kanan}}]{hayes2021replay}%
  \BibitemOpen
  \bibfield  {author} {\bibinfo {author} {\bibfnamefont {T.~L.}\ \bibnamefont {Hayes}}, \bibinfo {author} {\bibfnamefont {G.~P.}\ \bibnamefont {Krishnan}}, \bibinfo {author} {\bibfnamefont {M.}~\bibnamefont {Bazhenov}}, \bibinfo {author} {\bibfnamefont {H.~T.}\ \bibnamefont {Siegelmann}}, \bibinfo {author} {\bibfnamefont {T.~J.}\ \bibnamefont {Sejnowski}}, \ and\ \bibinfo {author} {\bibfnamefont {C.}~\bibnamefont {Kanan}},\ }\href@noop {} {\bibfield  {journal} {\bibinfo  {journal} {Neural Computation}\ }\textbf {\bibinfo {volume} {33}},\ \bibinfo {pages} {2908} (\bibinfo {year} {2021})}\BibitemShut {NoStop}%
\bibitem [{\citenamefont {Tadros}\ \emph {et~al.}(2022)\citenamefont {Tadros}, \citenamefont {Krishnan}, \citenamefont {Ramyaa},\ and\ \citenamefont {Bazhenov}}]{Tadros_NC2022}%
  \BibitemOpen
  \bibfield  {author} {\bibinfo {author} {\bibfnamefont {T.}~\bibnamefont {Tadros}}, \bibinfo {author} {\bibfnamefont {G.}~\bibnamefont {Krishnan}}, \bibinfo {author} {\bibfnamefont {R.}~\bibnamefont {Ramyaa}}, \ and\ \bibinfo {author} {\bibfnamefont {M.}~\bibnamefont {Bazhenov}},\ }\href@noop {} {\bibfield  {journal} {\bibinfo  {journal} {Nature Communications}\ }\textbf {\bibinfo {volume} {13}},\ \bibinfo {pages} {7742} (\bibinfo {year} {2022})}\BibitemShut {NoStop}%
\end{thebibliography}%

\end{document}